\documentclass[runningheads]{llncs}
\usepackage[T1]{fontenc}
\usepackage{color}
\usepackage{wrapfig}
\usepackage{multirow}
\usepackage{arydshln}
\usepackage{booktabs}
\usepackage{amsmath}
\usepackage{amssymb}
\usepackage{graphicx}
\usepackage{pifont}
\usepackage{tcolorbox}
\usepackage[misc]{ifsym}

\begin{document}
%
\title{Harnessing LLMs Explanations to Boost Surrogate Models in Tabular Data Classification}

\author{Ruxue Shi \and
Hengrui Gu \and
Xu Shen \and
Xin Wang \textsuperscript{(\Letter)}
}
\authorrunning{R.Shi et al.}
%
\institute{Jilin University, Changchun, China \\
\email{\{shirx24, guhr22, shenxu23\}@mails.jlu.edu.cn} \\ \email{xinwang@jlu.edu.cn}
}

\maketitle              
\begin{abstract}
Large Language Models (LLMs) have shown remarkable ability in solving complex tasks, making them a promising tool for enhancing tabular learning. However, existing LLM-based methods suffer from high resource requirements, suboptimal demonstration selection, and limited interpretability, which largely hinder their prediction performance and application in the real world. To overcome these problems, we propose a novel in-context learning framework for tabular prediction. The core idea is to leverage the explanations generated by LLMs to guide a smaller, locally deployable Surrogate Language Model (SLM) to make interpretable tabular predictions. Specifically, our framework mainly involves three stages: (i) Post Hoc Explanation Generation, where LLMs are utilized to generate explanations for question-answer pairs in candidate demonstrations, providing insights into the reasoning behind the answer. (ii) Post Hoc Explanation-Guided Demonstrations Selection, which utilizes explanations generated by LLMs to guide the process of demonstration selection from candidate demonstrations. (iii) Post Hoc Explanation-Guided Interpretable SLM Prediction, which utilizes the demonstrations obtained in step (ii) as in-context and merges corresponding explanations as rationales to improve the performance of SLM and guide the model to generate interpretable outputs. Experimental results highlight the framework's effectiveness, with an average accuracy improvement of 5.31\% across various tabular datasets in diverse domains.
\keywords{Large Language Models (LLMs)  \and Tabular data \and Surrogate Language Model (SLM).}
\end{abstract}
\section{Introduction}
\begin{figure}[t]
\centering
\includegraphics[width=1\textwidth]{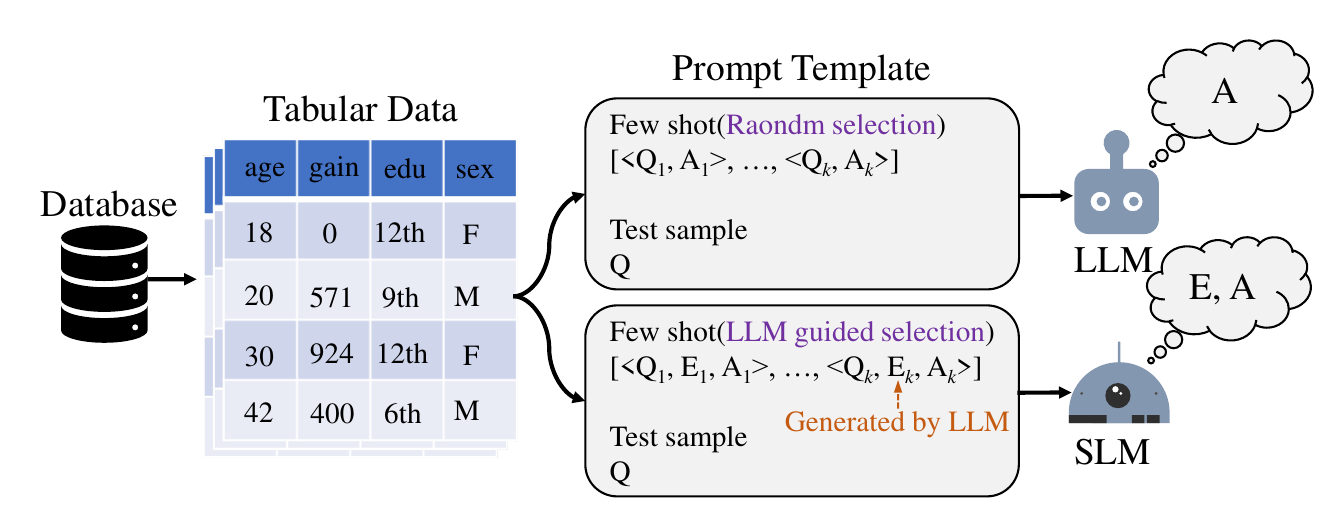}
\caption{\label{figure:Comparison}  Previous approaches to context learning with LLMs, as depicted on the top side of the figure, directly utilize tabular data to prompt the LLMs for predictions. This process necessitates frequent and extensive access to large language models (LLMs), resulting in substantial computational and efficiency demands. In contrast, our method depict at the bottom, presents a more efficient approach, which uses LLM generated explanations to guide the few shot demonstrations selection and the interpretable outputs of SLM}
\end{figure}

Recent advancements in model architecture and training methodologies have driven the remarkable progress and widespread adoption of Large Language Models (LLMs). By leveraging large-scale pre-training and fine-tuned instruction-based learning, LLMs have encoded extensive prior knowledge in their parameters. This has made them highly effective for various tasks, including dialogue generation \cite{finch2023leveraging} and code debugging \cite{fan2023automated}, without requiring task-specific training. Inspired by these successes, researchers have investigated applying LLMs to analyze tabular data, as depicted in Figure \ref{figure:Comparison} (top). Existing methods analyze tabular data sample-by-sample to uncover causally meaningful relationships between tabular features and their corresponding labels~\cite{dinh2022lift}. The ability of LLMs to generalize across different domains through such methods has set new benchmarks in tabular data analysis, particularly in data-scarce scenarios.
Despite these advancements, LLM-based approaches for tabular data face significant challenges:  \ding{182} The performance of LLMs in few-shot settings has shown to be sensitive to changes in prompts~\cite{zhangautomatic}. To optimize demonstrations for a given test sample, existing methods typically use representation similarity within the textual embedding space spanned by a pre-trained language model as the selection criterion. Due to the redundant nature of features in tabular data, this approach inadvertently incorporates irrelevant or spurious features into the calculated embeddings. This limits the representation of semantics introduced by causal tabular features, ultimately leading to a suboptimal selection scheme. \ding{183} Ensuring high-quality predictions often necessitates memory-intensive models or frequent API calls, both of which are computationally expensive for large-scale industrial applications. Additionally, most top-performing LLMs are accessible only through response-only APIs, which restrict downstream fine-tuning and limit their potential ability to be further aligned with specific domain requirements. \ding{184} Another critical limitation is the lack of interpretability in the predictions made by LLMs. While these black-box models excel at generating accurate results, their internal decision-making rule are often non-transparent, raising concerns about their reliability especially in critical applications such as medical diagnosis and autonomous driving.

To address these limitations, we propose a novel tabular learning framework that achieves significant improvements in both task performance and inference efficiency, offering a robust and effective solution to advance tabular learning tasks. Specifically, the framework introduces an additional warm-up stage prior to inference, where top-performing LLMs (e.g., ChatGPT) are tasked with generating feature attribution-based explanations on minimal task samples (candidate demon-
strations) that reflect their behavior patterns. This process identifies a subset of features that are causally meaningful to the decision-making process of these advanced LLMs. During the inference stage, the selected features are utilized to enable a more robust demonstration selection, focusing on retaining only relevant features and mitigating the impact of noisy ones. This approach facilitates precise and personalized demonstration choices that align closely with similar data patterns, thereby improving the quality of input prompts and enhancing overall performance (\textit{Challenge} \ding{182}). These high-quality demonstrations, combined with their corresponding predictive explanations, constitute an explanation-guided inference prompt. This approach enables the use of a computationally efficient small surrogate language model (SLM) as a substitute for the resource-intensive large-scale LLMs or costly commercial API calls. By leveraging the meticulously crafted prompt, the SLM can effectively learn the prediction patterns of top-performing LLMs and produce predictions economically without compromising task performance (\textit{Challenge} \ding{183}). Moreover, the generated explanations accompanying each prediction enhance interpretability by providing clear and transparent rationales for the model's outputs, fostering greater trust and understanding of its decision-making process (\textit{Challenge} \ding{184}).

The main contributions of this work are as follows:
\begin{itemize}
    \item We use LLMs as post hoc explanation generators to reduce the need for frequent API calls while still capturing key insights from the model.

    \item We introduce a novel framework that leverages LLMs-generated explanations to guide demonstration selection, and integrating these explanations as rationales to improve both classification performance and interpretability in surrogate language models.

    \item Our experiments show that the proposed framework leads to significant improvements in classification performance under few-shot learning conditions across multiple benchmark tabular datasets.
\end{itemize}
\section{Related Works}
\subsection{Classical Tabular Data}

Numerous machine learning methods have been developed for classification tasks on tabular data. For modeling linear relationships, logistic regression (LR) \cite{lavalley2008logistic} and generalized linear models (GLM) \cite{hastie2017generalized} are commonly used. For non-linear relationship modeling, tree-based models such as decision trees (DT) \cite{loh2011classification} and ensemble methods like XGBoost \cite{chen2016xgboost}, random forests \cite{breiman2001random}, CatBoost \cite{prokhorenkova2018catboost}, and LightGBM \cite{ke2017lightgbm} are widely applied. With the advancement of deep learning, there has been increasing interest in applying neural networks to tabular data. These methods can be categorized into four main types: 1) Standard Neural Networks: Examples include SNN \cite{klambauer2017self}, AutoInt \cite{song2019autoint}, and DCN V2 \cite{wang2021dcn}.
2) Hybrid Methods: These integrate decision trees with neural networks for end-to-end training, including NODE \cite{popov2019neural}, GrowNet \cite{badirli2020gradient}, TabNN \cite{ke2018tabnn}, and DeepGBM \cite{ke2019deepgbm}.
3) Transformer-Based Methods: These models leverage attention mechanisms to learn from features and data samples, as seen in TabNet \cite{arik2021tabnet}, TabTransformer \cite{huang2020tabtransformer}, and FT Transformer \cite{gorishniy2021revisiting}.
4) Representation Learning Methods: These methods use self-supervised and semi-supervised learning to extract meaningful information, with notable examples including VIME \cite{yoon2020vime}, SCARF \cite{bahri2021scarf}, SAINT \cite{somepalli2021saint}, and Recontab \cite{chen2023recontab}.

\subsection{In-Context Learning with Tabular Data}
In-context learning (ICL) offers an innovative approach by enabling language models to perform tasks based solely on input-output examples, without parameter updates or fine-tuning. This capability, first observed in GPT-3 and subsequent large language models (LLMs) \cite{liu2023pre}, allows models to generalize to new tasks by embedding examples directly into the input prompt. Often referred to as an "emergent ability" \cite{wei2022emergent}, this phenomenon has spurred significant research into understanding and enhancing ICL in models exceeding 100 billion parameters. To improve ICL, researchers have explored various strategies for enriching prompt content. For instance, the Chain-of-Thought (CoT) technique \cite{wei2022chain} incorporates human-annotated rationales, such as step-by-step task instructions, into prompts to enhance reasoning. In the context of tabular data, TABLET~\cite{slack2023tablet} integrates rule sets and prototypes from external classifiers to improve inference quality, while SPROUT~\cite{nam2023semi} combines unlabeled data with LLMs to extract transferable knowledge from tabular samples, reducing dependence on labeled data. Despite their success, these approaches often rely on passing all inference tasks through LLMs, which is computationally expensive and impractical for industrial-scale applications. Furthermore, they frequently overlook the critical importance of selecting optimal contextual demonstrations. This study aims to address these limitations by proposing a novel framework that moves away from a purely end-to-end reliance on LLMs inference. Instead, it leverages LLMs to infer the relationships between questions and answers directly. This relational insight is then used to guide the selection of demonstrations and generate rationales, enhancing both the in-context learning performance and the interpretability of surrogate language models.

\section{Method}
\begin{figure*}[t]
\centering
\includegraphics[width=1\textwidth]{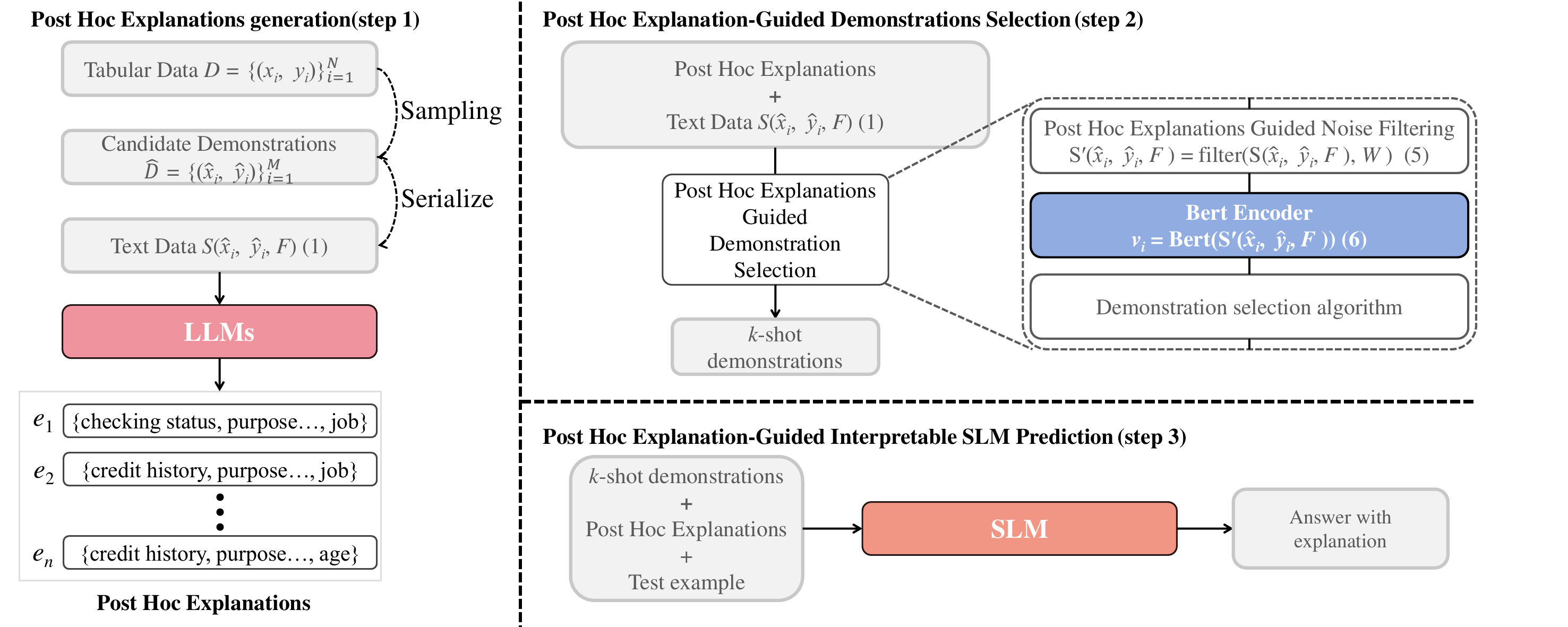}
\caption{\label{figure:method}Overview of our method. We begin by randomly selecting a small number of samples from the tabular dataset to form candidate demonstration sets. These samples are then converted into text and fed into an LLM, which generates post hoc explanations. These explanations are used to guide the selection of relevant demonstrations and to facilitate the generation of interpretable predictions by the SLM.}
\end{figure*}

Our framework enhances the performance and interpretability of the surrogate language model (SLM) by leveraging post hoc explanations derived from candidate demonstrations. These explanations guide the SLM in selecting the most relevant demonstrations, achieving superior downstream performance compared to directly prompt LLMs. As shown in Figure \ref{figure:method}, the process begins by transforming tabular data into a textual format suitable for processing by the language model. The LLMs then generate input-specific explanations, which are used to inform the selection of relevant demonstrations. Additionally, these explanations are integrated into the selected demonstrations as rationales, serving as a form of pre-feature engineering. By filtering out less significant features, this approach improves the accuracy and relevance of the selection process. Details of the problem formulation and each stage of the framework are discussed in the following sections.
\subsection{Preliminaries}
\textbf{Problem Formulation.} Consider a tabular dataset \( D = \{(\mathbf{x}_i, y_i)\}_{i=1}^N \), where \( N \) represents the total number of samples. Each sample \( \mathbf{x}_i \in \mathbf{X}\) is a \( d \)-dimensional vector, corresponding to \( d \) input features, with textual feature names denoted as \( F = \{f_j\}_{j=1}^d \). In the classification setting, the label \( y_i  \in \mathbf{Y} \) is drawn from a predefined set of classes. For \( k \)-shot learning experiments, a subset of \( k \) labeled samples, where \( k < N \), is randomly selected to train the model. The primary objective is to develop a model that can accurately predict the class label \( y_i  \) for each \( \mathbf{x}_i \) based on its feature representation, even with limited training samples.\\
\textbf{Tabular Data Serialization.} Tabular data serialization is a commonly used technique in tabular learning, converting tabular data into its corresponding textual form. Given a sample (\(\mathbf{\hat{x}}_i,\hat{y}_i\)) \( \in \hat{D}\), where candidate demonstrations set \(\hat{D}\) is obtained by randomly sampling \(M\) samples from \(D\), we define a transformation function, S, which converts each tabular sample into a question-answer format:

\begin{equation}
\begin{split}
\text{S}(\mathbf{\hat{x}}_i,\ \hat{y}_i,\ F) =\text{``Q:\  }
f_1 \ \text{ is } \ \hat{x}_{i1} \text{. }  \ldots  
f_d \ \text{ is } \ \hat{x}_{id} \text{. }
\, \text{A:\  } \hat{y}_i \text{”},
\end{split}
\end{equation}

\noindent where the vector \(\mathbf{\hat{x}}_i = (\hat{x}_{i1}, \dots, \hat{x}_{ij}, \dots, \hat{x}_{id})\) represents the feature values for the \(i\)-th sample, and \(\hat{x}_{ij}\) represents the value of the \(j\)-th column feature of the \(i\)-th sample. 

\subsection{Post Hoc Explanation Generation (step 1)}
\tcbset{
    colback=gray!10,       
    colframe=black,        
    boxrule=0.3mm,         
    arc=3mm,               
    fonttitle=\bfseries,   
    left=2mm, right=2mm,   
    top=1mm, bottom=1mm    
}
\begin{table}[t]
 \centering
  
\begin{tcolorbox}
    Task \\
    \textcolor[rgb]{ .929,  .49,  .192}{<Task descroption>} \\
    \\
    Candidate Demonstration \\
    \textcolor[rgb]{ .357,  .608,  .835}{<Q, A>} \\
     \\
    Post Hoc Explanation Generation Instruction \\
    \textcolor[rgb]{ .494,  .392,  .62}{Please provide \(n\) words in the question that are most important for obtaining the given answer.} \\
     \\
    Response Instruction \\
    \textcolor[rgb]{ .749,  .565,  0}{Note: The output format is as follows: \{'word', 'word', … , 'word', 'word'\}}
\end{tcolorbox}
\vspace{8pt}
\caption{\label{figure:instruction}Prompt for post hoc explanation generation. Text in orange provides the task description; Blue and green text represent the questions and answers in the candidate demonstration, respectively; purple text outlines post hoc explanation generation instruction; and yellow text details response instruction. }
\vspace{-15pt}
\label{tab:addlabel}%
\end{table}%

We seek to leverage the internal knowledge of LLMs to generate post hoc explanations for candidate demonstrations by using carefully designed instructions, as shown in the purple text in Table \ref{figure:instruction}. These instructions are crafted to clarify the reasoning behind each question-answer (Q-A) pair, enhancing the interpretability and performance of the surrogate language model (SLM). Specifically, the LLMs identify and select the \( n \) most salient features from \( Q \) that are most relevant to \( A \). These features serve as the post hoc explanations for the candidate demonstration, highlighting the critical aspects of the input-output relationship. This approach not only improves transparency but also ensures that the model focuses on meaningful and causal correlations, avoiding irrelevant or spurious associations.

To ensure reply consistency and usability, we guide the LLMs to structure their responses according to predefined instructions, illustrated in the yellow text in Table \ref{figure:instruction}. These prompts specify the format and required number of post hoc explanations, enabling straightforward parsing and application of the generated outputs. The formal process for generating post hoc explanations for the \( i \)-th candidate demonstration is defined as follows:

\begin{equation}
e_i = \{w_{1}, w_{2}, \dots, w_{n}\} = \text{LLMs}(\text{S}(\mathbf{\hat{x}}_i,\ \hat{y}_i,\ F),\ I),
\end{equation}

\noindent where \( e_i \) represents the set of salient features, \( \mathbf{\hat{x}}_i \) and \( \hat{y}_i \) are the input and label of the \( i \)-th candidate demonstration, \( F \) is the set of feature names, and \( I \) represents the instructional prompts used to guide the LLMs. This structured approach ensures clarity, task alignment, and improved model performance.

\subsection{Post Hoc Explanation-Guided Demonstrations Selection (step 2)}

We propose an innovative approach for demonstration selection, which leverages post hoc explanations generated by LLMs to guide the identification of the most relevant and informative features, enabling improved in-context learning performance. The objective is to prioritize features that contribute most significantly to the quality of demonstrations. To achieve this, we compute the importance of each feature, \( g(w_j) \), using the following function:

\begin{equation}
    g(w_j) = \frac{1}{n \times M} \sum_{i=1}^{M} \mathbf{1}_{\{w_j \in e_i\}},
\end{equation}

\noindent where \( M \) is the total number of candidate demonstrations, and \( \mathbf{1}_{\{w_j \in e_i\}} \) is an indicator function that equals 1 if the feature \( w_j \) appears in the explanation \( e_i \). Based on this, we define the set \( W \), which consists of the top \( q \) most important features ranked by their importance:

\begin{equation}
    W = \{ w_i \mid w_i \in \text{Top}_{q}((w_i, g(w_i), p)) \},
\end{equation}

\noindent where \( p \) serves as the feature importance threshold, guiding the selection of significant features. The function \(\text{Top}_{q}((w_i, g(w_i), p))\) identifies the top \( q \) most important features by evaluating their importance levels against this threshold. These selected features are then used to filter the candidate demonstrations, this process is mathematically represented as:

\begin{equation}
    \text{S}'(\mathbf{\hat{x}}_i, \hat{y}_i, F) = \text{filter}(\text{S}(\mathbf{\hat{x}}_i, \hat{y}_i, F), W),
\end{equation}

\noindent where the function \( \text{filter(\(\cdot\))} \) removes any sentences in \( \text{S}(\mathbf{\hat{x}}_i, \hat{y}_i, F) \) that do not contain features from \( W \). To encode the filtered demonstrations, we use Bert:

\begin{equation}
    v_i = \text{Bert}(\text{S}'(\mathbf{\hat{x}}_i, \hat{y}_i, F)),
\end{equation}

\noindent where \(v_i \in \mathbb{R}^u\) represent the vectorized representation of the \(i\)-th filtered demonstration and \(u\) is the hidden size of Bert. To evaluate the importance score \(s_i\) for \(i\)-th filtered candidate demonstration, we employed diversity-based algorithms (Cluster-Based Methods) and similarity-based algorithms, including Cosine Similarity, Euclidean Distance, and Manhattan Distance. The impact of these methods across various datasets is analyzed detailedly in Section 4.7.
\begin{itemize}

    \item \textbf{Cluster-Based Methods.} To obtain demonstrations of diversity, based on the candidate demonstrations, we run a K-Means clustering to generate the importance score \(s_i\).
    \begin{equation}
     s_i = -\min_{\mathbf{C} \in \mathbb{R}^{ d \times m}} 
    \left\| v_i - \mathbf{C} \right\|_2^2,
    \end{equation}
    where \(m\) is the number of centroids, and \(\mathbf{C}\) is the centroid matrix.
    
    \item \textbf{Cosine Similarity-Based Methods.} We use cosine similarity to obtain a demonstration with higher semantic similarity to the test sample \(x_t\), the semantic similarity score between the \(i\)-th candidate demonstration \( v_i \) and the embedded test sample \( v_t \) is calculated using the following formula:
    \begin{equation}
     s_i = v_t \odot v_i,
    \end{equation}
    where \(\odot\) represent element multiplication.

    \item \textbf{Euclidean Distance-Based Methods.} We also considered the influence of physical distance on similarity scores, and the calculation formula is as follows:
    \begin{equation}
     s_i = - \left\| v_t - v_i \right\|_2^2
    \end{equation}

    \item \textbf{Manhattan Distance-Based Methods.} In order to reduce the influence of outliers in physical distance, we also use Manhattan distance to calculate the importance score \(s_i\):
    \begin{equation}
     s_i = - \left\| v_t - v_i \right\|
    \end{equation}
    
\end{itemize}

The final \( k \)-shot demonstrations are selected based on the score \(s_i\):

\begin{equation}
    k\text{-shot} = \{ \text{S}(\mathbf{\hat{x}}_i, \hat{y}_i, F) \mid \text{S}(\mathbf{\hat{x}}_i, \hat{y}_i, F) \in \text{Top}_k(\text{S}(\mathbf{\hat{x}}_i, \hat{y}_i, F), s_i) \}
\end{equation}

\noindent Our framework systematically combines feature importance, explanation-guided filtering, and scoring algorithms to optimize the selection of demonstrations for in-context learning, improving both performance and interpretability.

\subsection{Post Hoc Explanation-Guided Interpretable SLM Prediction (step 3)}

To integrate post hoc explanations into the learning process, the \( k \)-shot demonstrations are enhanced by appending their corresponding explanations, as defined below:

\begin{equation}
    k\text{-shot}' = \{ \text{S}(\mathbf{\hat{x}}_i, \hat{y}_i, F) + e_i \mid \text{S}(\mathbf{\hat{x}}_i, \hat{y}_i, F) \in k\text{-shot} \}
\end{equation}

\noindent Using these enriched demonstrations, the model generates predictions for a test sample by applying the SLM with the updated \( k \)-shot demonstrations:

\begin{equation}
    reply = \text{SLM}(k\text{-shot}', S(x_t, \text{-}, F)),
\end{equation}

\noindent where the model's reply consists of both an explanation (\( e \)) and a predicted answer (\( A \)). A parser is then employed to extract these components:

\begin{equation}
  e, A = \mathrm{Parser}(reply)  
\end{equation}

\noindent This process ensures that the model not only provides accurate predictions but also includes clear, interpretable explanations, enhancing the transparency and reliability of the predictions.







\section{Experimental Evaluation}






In this section, we seek to address the following research questions:

\begin{itemize}

   \item (\textbf{RQ1.}) How does our proposed method compare to baseline approaches in a few-shot learning setting?  

   \item (\textbf{RQ2.}) How well does our method balance performance with reduced API usage compared to ChatGPT?  

   \item (\textbf{RQ3.}) What is the contribution of each individual component of our method to the overall performance?  

   \item (\textbf{RQ4.}) How do different selection algorithms impact model performance? Specifically, does the Post Hoc Explanation Guided Demonstrations Selection approach improve the model's ability to choose the most appropriate examples?

\end{itemize}

\subsection{Datasets}

\begin{wraptable}[9]{r}{0.6\textwidth}
  \centering
  \vspace{-8mm}

  \resizebox{0.58\textwidth}{!}{
    \begin{tabular}{p{14em}cccc}
    \toprule
    Dataset & \multicolumn{1}{p{3em}}{\centering bank} & \multicolumn{1}{p{3em}}{\centering creditg} & \multicolumn{1}{p{3em}}{\centering heart} & \multicolumn{1}{p{3em}}{\centering income} \\
    \midrule
    \# of train samples & 40689 & 900   & 826   & 43958 \\
    \# of test samples & 4522  & 100   & 92    & 4884 \\
    \# of candidate demonstrations & 100   & 100   & 100   & 100 \\
    \# of features & 16    & 20    & 11    & 12 \\
    \bottomrule
    \end{tabular}%
    }
    \vspace{8pt}
   \caption{Basic information of each dataset used in our experiments.}
  \label{tab:addlabel}%
\end{wraptable}

We use four datasets for binary classification tasks in our experiments:  The Bank dataset~\cite{moro2014data} for predicting whether a customer will subscribe to a term deposit is a dataset about marketing activities of Portuguese banking institutions, containing 45211 records and 16 features; The creditg dataset~\cite{kadra2021well} contains 1000 records and 20 features, suitable for predicting credit ratings; The Income dataset~\cite{asuncion2007uci} has 48842 records and 14 features, focusing on the classification of individual income levels; The Heart dataset\footnote{\url{kaggle.com/datasets/fedesoriano/heart-failure-prediction}} has 918 records and 11 features, focusing on identifying and predicting patients with heart disease.

Each dataset was partitioned into training and testing sets with a 9:1 split. To reduce computational complexity and minimize the number of queries to the LLMs, we randomly selected 100 samples from the training set as candidate demonstrations. Table \ref{tab:addlabel} summarizes the key details of these datasets, including the number of training and test samples, as well as the number of features.

\subsection{Experimental Setting}

Our framework employs GPT-3.5\footnote{\url{https://openai.com/blog/chatgpt}} as the backbone for the LLMs. For the LLMs inference, the temperature is set to 0.3 to balance creativity and consistency, while the top-p value remains at the default of 1, as provided by the API settings. We configure the number of post hoc explanations \(n\) to 5 and set the filtering threshold \(p\) for filtering noisy sentences from the text data based on feature importance at 0.85. For encoding text data, we use SentenceBERT with a hidden dimension of 128. Additionally, the Llama2-7B model~\cite{touvron2023llama} is used as SLM for tabular classification, leveraging in-context learning with 4-shot demonstrations for better generalization across diverse scenarios. In replicating baseline models, we include traditional machine learning methods such as Logistic Regression (LR), LightGBM, and Random Forest. The hyperparameters for these models were optimized using Grid Search combined with \(K\)-fold cross-validation. The value of \(K\) was set to either 2 or 4, ensuring that the training set contained at least one example from each class.
\subsection{Baseline Methods}

We compare our framework against eight baseline models, categorized into three groups. The first set includes conventional supervised learning methods for tabular data, specifically: (1) Logistic Regression (LR), (2) LightGBM~\cite{ke2017lightgbm}, and (3) Random Forest (RF)~\cite{ho1995random}. These models represent widely used, classical approaches for structured data classification. The second group comprises models that use pretraining followed by fine-tuning on specific tasks, including: (4) SCARF~\cite{bahri2022scarf}, (5) STUNT~\cite{nam2023stunt}, and (6) TabPFN~\cite{hollmann2022tabpfn}. The final group consists of LLMs-based methods, namely: (7) In-context Learning(AO)~\cite{wei2022emergent} and (8) TABLET~\cite{slack2023tablet}. In-context learning (AO) embeds few-shot training examples directly within the input prompt, without any parameter tuning. TABLET enhances in-context learning by incorporating additional information, such as rule sets and prototypes from an external classifier, into the prompt to improve inference quality. It is worth noting that, for a fair comparison, we replicated both AO and TABLET using the same Llama2-7B model as the backbone to ensure consistency across the experiments.

\subsection{Results and Analysis(RQ1)}	
\begin{table*}[t]
  \centering

    \begin{tabular}{p{7.665em}ccccc}
    \toprule
     & \multicolumn{5}{p{21.605em}}{\centering Dataset} \\
\cmidrule{2-6}  \multicolumn{1}{c}{\centering Method} & \multicolumn{1}{p{3.165em}}{\centering bank} & \multicolumn{1}{p{3.165em}}{\centering creditg} & \multicolumn{1}{p{3.89em}}{\centering heart} & \multicolumn{1}{p{4.72em}}{\centering income} & \multicolumn{1}{p{6.665em}}{\centering All Tasks (avg)} \\
    \midrule
    \centering LR    & 53.00\(_{\text{22.00}}\) & 51.00\(_{\text{13.00}}\) & 68.00\(_{\text{6.00}}\) & 67.00\(_{\text{4.00}}\) & 59.75 \\
    \centering RF    & 48.00\(_{\text{10.00}}\) & 55.00\(_{\text{4.00}}\) & 73.00\(_{\text{3.00}}\) & 61.00\(_{\text{6.00}}\) & 59.25 \\ 
    \centering LightGBM & 51.00\(_{\text{3.00}}\) & 42.00\(_{\text{17.00}}\) & 59.00\(_{\text{1.00}}\) & 64.00\(_{\text{6.00}}\) & 54.00 \\ \hdashline
    \centering SCARF & 55.00\(_{\text{7.00}}\) & 54.00\(_{\text{13.00}}\) & 68.00\(_{\text{4.00}}\) & 61.00\(_{\text{9.00}}\) & 59.50 \\
    \centering STUNT & 54.67\(_{\text{6.65}}\) & \textbf{64.33}\(_{\text{0.94}}\) & 53.26\(_{\text{4.94}}\) & 66.33\(_{\text{6.94}}\) & 59.65 \\ 
    \centering TabPFN & 49.00\(_{\text{6.00}}\) & 58.00\(_{\text{4.00}}\) & 59.00\(_{\text{4.00}}\) & 67.00\(_{\text{2.00}}\) & 58.25 \\ \hdashline
    \centering AO    & 35.23\(_{\text{0.88}}\) & 62.79\(_{\text{1.26}}\) & 60.59\(_{\text{3.29}}\) & 57.13\(_{\text{3.49}}\) & 53.94 \\
    \centering TABLET & 21.00\(_{\text{3.55}}\) & 38.40\(_{\text{3.45}}\) & 60.30\(_{\text{2.95}}\) & 61.60\(_{\text{3.95}}\) & 45.33 \\
    \centering ours  & \textbf{55.26}\(_{\text{0.88}}\) & \textbf{64.33}\(_{\text{2.19}}\) & \textbf{73.06}\(_{\text{1.81}}\) & \textbf{67.58}\(_{\text{2.00}}\) & \textbf{65.06} \\
    \bottomrule
    \end{tabular}%
    \vspace{10pt}
    \caption{4-shot test accuracy results across 4 datasets. All the remaining tables in this paper follow these setups to avoid clutter: the metric values are averaged over 3 random seeds and standard deviations are given as subscripts; the All Tasks (avg) column reports the average accuracy across all datasets; Top results for each dataset are in bold.}
  \label{tab:addlabel}%
    
\end{table*}%

Table \ref{tab:addlabel} presents the 4-shot test accuracy results across four datasets. Our proposed framework consistently outperforms all baselines across the datasets, achieving a more than 5\% performance improvement for all settings. Notably, conventional supervised learning approaches (Logistic Regression (LR), Random Forest (RF), and LightGBM) demonstrate mixed results, such as Logistic Regression performs well on the income and heart dataset but fails to maintain competitiveness across other datasets. Random Forest performs similarly, excelling on the heart dataset but showing lower accuracy on the bank and creditg datasets. Pre-trained models (SCARF, STUNT, and TabPFN) have more competitive performance. Compared to conventional supervised learning approaches, pre-trained models benefit from incorporating prior knowledge related to downstream tasks during pre-train, leading to more stable performance across different datasets with minimal fluctuations. In contrast, the in-context learning models AO and TABLET struggle to match the performance of the other approaches. This outcome is reasonable, as the utilization of the less powerful Llama2-7B model significantly constrains the overall effectiveness of these methods. In summary, our framework demonstrates a clear advantage over both conventional supervised learning models and advanced methods utilizing pretraining or in-context learning. our framework utilizes the explanations of LLMs to guide SLM in inference, contributing to its superior performance across all datasets. These results validate the effectiveness of leveraging our method in few-shot tabular classification.

\subsection{Backbone Model Result(RQ2)}
\begin{table*}[t]
  \centering

    \begin{tabular}{p{7.89em}cc}
    \toprule
    Method & \multicolumn{1}{p{6.78em}}{\centering ACC (avg)} & API Usage (avg)\\
    \midrule
    AO(ChatGPT) & 59.90 & 2400\\
    ours(Llama2-7B) & 65.06 & 100\\ 
    ours(ChatGPT) & 68.79 & 2500\\
    \bottomrule
    \end{tabular}%
    \vspace{10pt}
      \caption{
Comparative analysis of trade-off between model performance and API usage
}
\vspace{-15pt}
\label{tab:backbone_model}
\end{table*}%

Table \ref{tab:backbone_model} presents the average accuracy and corresponding average API usage for ChatGPT across all tasks, comparing our proposed methods with AO(ChatGPT). The results demonstrate that our approach utilizing the locally deployable Llama2-7B model significantly reduces API usage while achieving higher average accuracy (6.18\% improvement over AO(ChatGPT)). Furthermore, when directly using ChatGPT as the proxy model, our method achieves a substantial performance gain (8.89\% improvement) with only a modest increase in API usage (2400->2500). This highlights the flexibility of our approach in balancing model performance and resource efficiency based on task requirements.
 
\subsection{Ablation Study(RQ3)}
\begin{table*}[t]
  \centering

    \begin{tabular}{p{4.89em}ccccc}
    \toprule
    Method & \multicolumn{1}{p{4em}}{\centering bank} & \multicolumn{1}{p{4em}}{\centering creditg} & \multicolumn{1}{p{4em}}{\centering heart} & \multicolumn{1}{p{4em}}{\centering income} & \multicolumn{1}{p{6.78em}}{\centering All Tasks (avg)} \\
    \midrule
    ours  & \textbf{55.26}\(_{\text{0.88}}\) & \textbf{64.33}\(_{\text{2.19}}\) & \textbf{73.06}\(_{\text{1.81}}\) & \textbf{67.58}\(_{\text{2.00}}\) & \textbf{65.06} \\
    \midrule
    -post hoc & 40.31\(_{\text{2.68}}\) & 54.36\(_{\text{1.61}}\) & 67.83\(_{\text{0.85}}\) & 66.04\(_{\text{1.34}}\) & 57.14\\
    -select & 45.15\(_{\text{2.61}}\) & 63.53\(_{\text{2.76}}\) & 56.60\(_{\text{3.38}}\) & 62.45\(_{\text{2.05}}\) & 56.93 \\
    -both(AO) & 35.23\(_{\text{0.88}}\) & 62.79\(_{\text{1.26}}\) & 60.59\(_{\text{3.29}}\) & 57.13\(_{\text{3.49}}\) & 53.94 \\
    \bottomrule
    \end{tabular}%
    \vspace{10pt}
      \caption{
Ablation of model performance
}
\vspace{-8pt}
\label{tab:Ablation_result}
\end{table*}%

We evaluate the contributions of key components to the overall performance through one-by-one ablation studies, focusing on the following: (1) \(-\text{post hoc}\): excluding the process of adding the post hoc explanations to the \(k\)-shot demonstrations; (2) \(-\text{select}\): removing the process of Post Hoc Explanation-Guided Demonstrations Selection, and (3) \(-\text{both (AO)}\): omitting all components.

Table \ref{tab:Ablation_result} summarizes the change in the acc from ablations as well as the average performance across all datasets. The results show that removing any of the key components leads to a noticeable drop in performance. When post hoc explanations are removed (\text{-post hoc}), the performance decreases across all datasets, with the bank dataset experiencing the most significant drop in accuracy (from 55.26\% to 40.31\%). This decline highlights the contribution of post hoc explanations in providing context and improving interpretability for the surrogate language model, thus aiding in accurate classification. Similarly, excluding the demonstration selection component (\text{-select}) results in a noticeable decrease in performance. This suggests that the selection of relevant demonstrations is critical for guiding the model effectively, as it enables the model to leverage representative examples that are closely aligned with the target samples. The combined ablation (\text{-both (AO)}) yields the lowest average performance (53.94\%), demonstrating the compounded effect of removing both components. The combined ablation performs significantly worse than the full model, indicating that both post hoc explanations and guided demonstration selection are essential for maximizing model accuracy on tabular data classification tasks. In summary, these ablation results confirm that each component contributes uniquely to the model's success, and their combination is essential to achieving robust, high-performing results across varied datasets. The findings further validate the efficacy of our proposed approach in leveraging explanations and optimized demonstration selection to enhance few-shot learning for tabular data classification.
\subsection{Impact of Few-shot demonstrations Selection algorithm(RQ4)}	
\begin{figure}[t]
\centering
\resizebox{0.8\textwidth}{!}{
\includegraphics[width=1\textwidth]{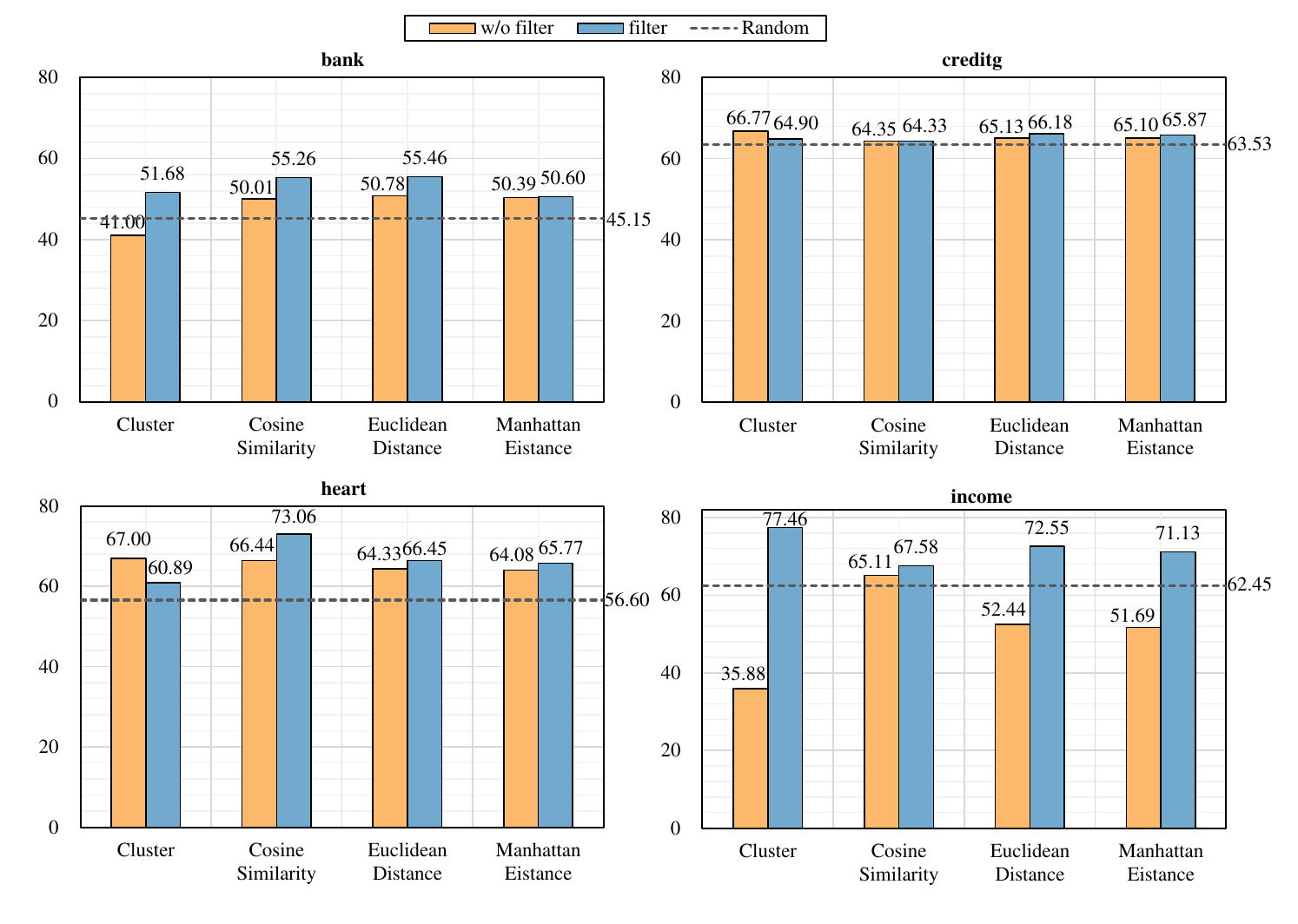}
}
\caption{\label{figure:selection_fun}Comparison of different demonstration selection algorithms and baselines (random selection) on four datasets.}
\vspace{-8pt}
\end{figure}

Figure \ref{figure:selection_fun} compares the performance of our framework under different few-shot demonstration selection algorithms across four tasks on tabular datasets, using the random demonstration selection algorithm as the baseline. The results reveal that for diversity-based selection algorithms (clustering) and distance-based similarity algorithms (Euclidean distance and Manhattan distance), the performance of the unfiltered(w/o filter) approach is occasionally inferior to random selection on certain datasets. However, with the integration of filtering(filter), all these algorithms outperform random selection consistently across all datasets. This demonstrates the critical role of filtering in mitigating noise during the demonstration selection process, enabling the selection of optimal demonstrates to guide the SLM. Additionally, the cosine similarity-based selection algorithm outperforms random selection across all datasets, regardless of the presence of filtering. This consistent superiority underscores the algorithm's robustness and suitability for tabular data, making it a reliable choice for demonstration selection in few-shot learning settings. These findings highlight the benefits of filtering and the effectiveness of cosine similarity in enhancing the quality of few-shot demonstrations for downstream performance.
\subsection{Hyper-parameter analysis}
\begin{figure}[t]
\centering
\includegraphics[width=0.7\textwidth]{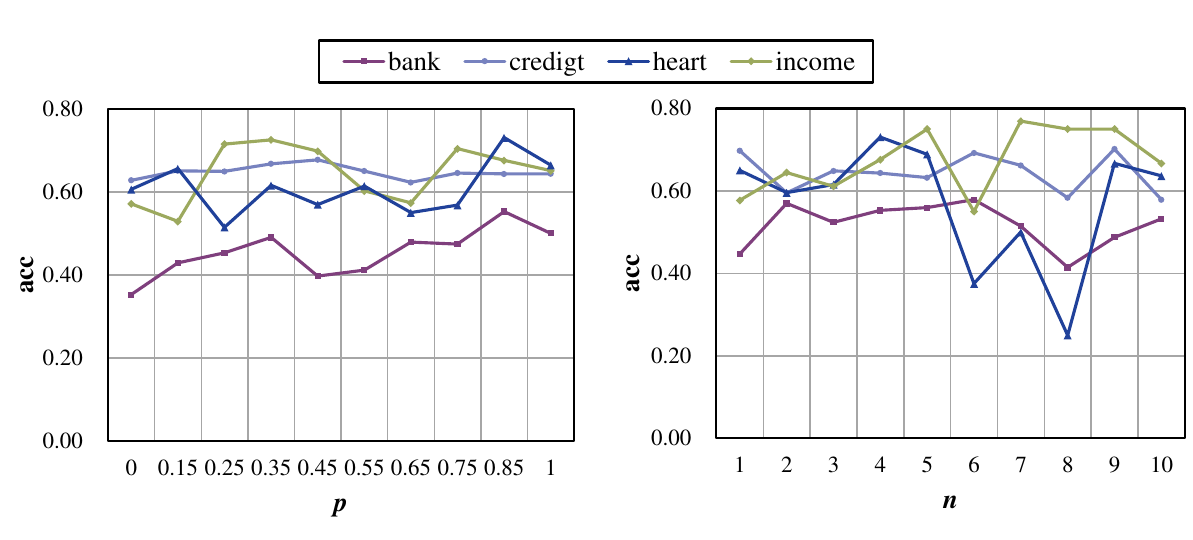}
\caption{\label{figure:Hyper-parameter p}Effect of filter threshold \(p\) and the number of post hoc explanations \(n\). The accuracy (acc) across all datasets is reported.}
\end{figure}


Figure \ref{figure:Hyper-parameter p} illustrates the performance of our model across four datasets, under two different hyper-parameters: \( p \) (left) and \( n \) (right). In the left Figure, acc values are displayed as a function of the parameter \( p \), which ranges from 0 to 1. The results indicate a relatively stable acc across all datasets, with minor fluctuations as \( p \) varies. Notably, the income and creditg datasets maintain a higher and more consistent acc performance compared to the bank and heart datasets. The heart dataset shows greater variability, with peaks and troughs as \( p \) increases, suggesting sensitivity to changes in this parameter.

The right plot presents acc values as a function of parameter \( n \), which takes integer values from 1 to 10. where the income and creditg datasets again demonstrate relatively stable acc scores. Conversely, the heart dataset exhibits substantial variability, particularly a marked drop in acc at \( n = 6 \), before recovering at higher values. This variability may indicate that the heart dataset is more sensitive to the selection or quantity of examples (parameter \( n \)) than the other datasets.

Overall, these results suggest that the choice of parameter \( n \) may have a significant impact on model performance, particularly for datasets like heart. In contrast, the parameter \( p \) appears to have a more modest effect on acc across datasets, though some variability is noted. The income and creditg datasets appear to be more robust to changes in both \( p \) and \( n \), consistently achieving higher acc values, while the bank and heart datasets show more sensitivity, highlighting the importance of parameter tuning for optimal performance on these datasets.
\vspace{-15pt}
\section{Conclusion}
\vspace{-10pt}
This paper proposes an in-context learning framework leveraging the priori knowledge of Large Language Models (LLMs) to guide a SLM for tabular learning. By using LLMs as post hoc explanation generators, we address challenges like efficient demonstration selection and reduce dependency on resource-intensive models, achieving a practical, interpretable, and cost-efficient learning process. Experimental results showed a 5.31\% accuracy improvement across tabular datasets, demonstrating the framework's effectiveness, especially in data-limited environments. The framework provides a scalable balance between performance and operational cost. Future work will focus on extending these methods to other data types and refining demonstration selection for broader applications.
\vspace{-15pt}
\subsubsection{\ackname} This work was supported by a grant from the National Natural Science Foundation of China under grants (No.62372211,  62272191), and the International Science and Technology Cooperation Program of Jilin Province (No.20230402076GH, No. 20240402067GH), and the Science and Technology Development Program of Jilin Province (No. 20220201153GX).

\bibliography{dasfaa2025}
\bibliographystyle{splncs04}

\end{document}